\documentclass[letterpaper,11pt]{article}
\usepackage[margin=1in,letterpaper]{geometry} 

\usepackage{subfigure} 
\usepackage{natbib}
\usepackage[utf8]{inputenc} 
\usepackage[T1]{fontenc}    
\usepackage[colorlinks,citecolor=blue]{hyperref}  
\usepackage{url}            
\usepackage{booktabs}       
\usepackage{amsfonts}  
\usepackage{times}
\usepackage{algorithmic}
\usepackage{algorithm}
\usepackage{nicefrac}       
\usepackage{microtype}      
\usepackage{amsthm}
\usepackage{float}
\usepackage{graphicx}
\usepackage{array}
\usepackage{multirow}
\usepackage{ragged2e}
\usepackage{color}
\usepackage{arydshln}
\usepackage{cite}
\usepackage{amsmath}
\usepackage{tabularray}
\usepackage{wrapfig}
\usepackage{amssymb}
\usepackage{todonotes}
\usepackage{xcolor}
\usepackage{xspace}
\usepackage{colortbl}

\usepackage{multicol}
\usepackage{enumerate}
\usepackage{xspace}

\newif{\ifhidecomments}
\hidecommentsfalse
\ifhidecomments
\newcommand{\wjdd}[1]{\todo[linecolor=cyan,backgroundcolor=cyan!25,bordercolor=cyan,size=\scriptsize]{}
\newcommand{\wjd}[1]{{\color{cyan}{}
\else
\newcommand{\wjdd}[1]{\todo[linecolor=cyan,backgroundcolor=cyan!25,bordercolor=cyan,size=\scriptsize]{(Jindong): #1}}
\newcommand{\wjd}[1]{{\color{cyan}{[(Jindong): #1]}}}
\fi

\newcommand{\cl}[1]{[\textcolor{red}{cl: #1}]}
\newcommand{\prompt}[1]{{\small \ttfamily #1}}

\newcommand{\method}{\textsc{MSTemp}\xspace}
\newcommand{\llms}{LLMs\xspace}

\title{Meta Semantic Template for Evaluation of Large Language Models}

\author{Yachuan Liu$^1$\thanks{Work done during internship at Microsoft Research Asia. Contact: yachuan@umich.edu.}, Liang Chen$^2$, Jindong Wang$^3$\thanks{Corresponding author: Jindong Wang (jindong.wang@microsoft.com). Updated work at: \url{https://llm-eval.github.io/}.}, Qiaozhu Mei$^1$, Xing Xie$^3$ \\
  $^1$University of Michigan \quad $^2$The Chinese University of Hong Kong \quad $^3$Microsoft Research}

\begin{document}
\maketitle
\begin{abstract}
Do large language models (LLMs) genuinely understand the semantics of the language, or just memorize the training data? The recent concern on potential data contamination of LLMs has raised awareness of the community to conduct research on LLMs evaluation. In this paper, we propose \method, an approach that creates meta semantic templates to evaluate the semantic understanding ability of LLMs. The core of \method is not to perform evaluation directly on existing benchmark datasets, but to \emph{generate} new out-of-distribution (OOD) evaluation sets using existing datasets as \emph{seeds}. Specifically, for a given sentence, \method leverages another language model to generate new samples while preserving its semantics. The new samples are called semantic templates to the original sentence. Then, \method generates evaluation samples via sentence parsing and random word replacement on the semantic templates. \method is highly flexible, dynamic, and cost-effective. Our initial experiments show that \method-generated samples can significantly reduce the performance of LLMs using existing datasets as seeds. We hope this initial work can shed light on future research of LLMs evaluation.

\end{abstract}

\section{Introduction}

Recently, large language models (LLMs) have achieved considerable performance in various applications \citep{chang2023survey,bubeck2023sparks}.
The so-called ``foundation model'' \citep{bommasani2021opportunities} tackles the downstream tasks through in-context learning \citep{brown2020language} by taking examples from the prompts.
However, the great performance of LLMs is being challenged these days, with concerns about potential data contamination since they are typically trained on public or open-source datasets \citep{zevcevic2023causal,carlini2022quantifying,biderman2023emergent,berglund2023reversal}.
For example, \citet{berglund2023reversal} found that LLMs trained in the format of ``A is B'' failed to generalize to evaluation samples in the form of ``B is A'', even if they re-trained the model in the format of ``B is A''.
Their main claim is that LLMs might fail to learn the causality of entity relations even present training data in the same format.
Similarly, \citet{robust2023dialog} reported that the order of input sentences significantly influences the quality of LLMs' responses, even when the models have been fine-tuned with various sentence orders.
In a nutshell, it requires to design of new evaluation protocols to test the ``true intelligence'' of \llms.

This paper aims to tackle the possibly contaminated evaluation of \llms by \emph{creating} dynamic and diverse samples.
Our main idea is to not rely on existing datasets or manual collection of new data \citep{ma2021dynaboard,kiela2021dynabench,thrush2022dynatask} as evaluation sets since they could be easily memorized by \llms.
However, the creation of new samples is not trivial. An ideal sample generation method should possess several characteristics. 
First, the new samples should be natural and fluent as they are assumed to test the language abilities of \llms.
Second, in order to generate infinite samples, existing datasets should be fully leveraged.
Third, the generation algorithm should have the ability to generate samples at different difficulty levels since the capability of \llms is stronger.
An ideal state is that we treat existing datasets as ``seeds'' and then our new evaluation protocols can generate new samples based on these seeds.

\begin{figure}[t!]
    \centering
    \includegraphics[width=1\linewidth]{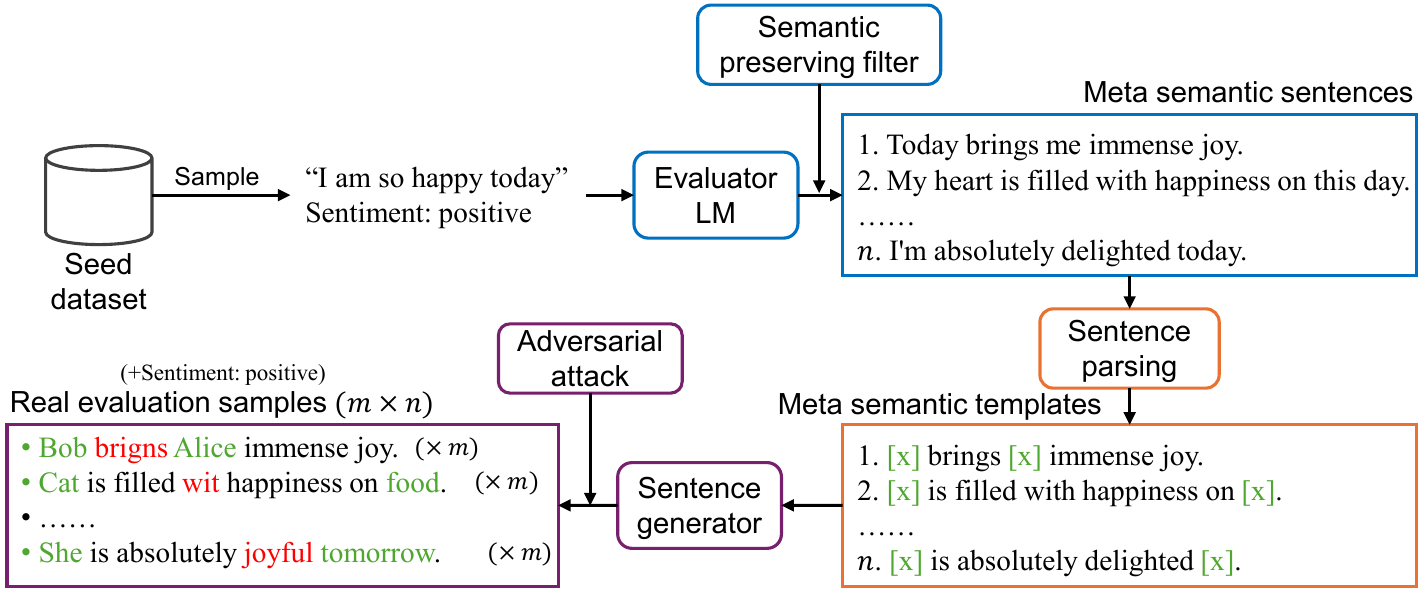}
    \vspace{-.1in}
    \caption{The procedure of \method. It consists of three parts: meta semantic sentence generation (blue), meta semantic templates generation (orange), and real evaluation samples generation (purple).}
    \label{fig-main}
    \vspace{-.15in}
\end{figure}

In this work, we propose \textbf{\method}, an evaluation protocol that creates \emph{meta semantic templates} to generate new testing samples.
The core of \method is to leverage language model $A$ to generate evaluation samples to test language model $B$ via creating meta semantic templates on an existing dataset $D$.
Specifically, a sentence $x \in D$ can be rephrased by $A$ to generate new sentences $[s_1, s_2,\cdots,s_n]$, which we call the meta semantic templates.
Then, real evaluation samples are generated based on these templates through a sentence parsing procedure to randomly replace different modules of the templates.
Using model $A$ can help to maintain the naturalness of the generated samples and most importantly, preserve the semantics of the original seed sentence.
We further introduce a semantic preserving filter to control the semantic difference between the original sample $x$ and the generated template $s$.
By design, \method is able to reduce the possibility of data contamination due to many choices of the evaluator language model $A$ and the seed dataset $D$ (imagine the efforts of generating training samples by replacing many $A$s and $D$s if someone wants to cheat).
\figurename~\ref{fig-main} illustrates the main procedure of \method.

The \method framework remains general and flexible to many language tasks.
For any given seed dataset, \method can generate new evaluation samples by controlling the number of templates ($n$ in the figure) and the real generated samples ($m$).
Moreover, it can naturally control the difficulty of generated samples via the adversarial attack module.
\citet{zhu2023promptbench} and \citet{wang2023robustness} showed that \llms are sensitive to adversarial attacks, among which the word-level attacks, i.e., replacing words in a sentence with their synonyms, remains the most successful threat.
Other attacks such as typos exhibit less success rate, compared to word-level attacks.
Therefore, the adversarial attack module in \method can easily control the complexity of the generated samples by adding different types of adversarial attacks.
The adversarial texts are marked with red in \figurename~\ref{fig-main}.

From a broader perspective, \method actually attempts to generate \emph{out-of-distribution} (OOD) samples for evaluation \llms, which remains a major threat to current \llms \citep{yang2022glue,arora2021types}.
While the predominant proprietary and some open-sourced \llms are trained on unknown sources of data, it becomes impossible to define what are proper OOD samples for them.
Therefore, \method can be seen as a manageable way to perform OOD evaluation of large language models in a straightforward manner.

This is a work-in-process paper and we are planning to conduct extensive experiments to verify the effectiveness of \method in the future.
Our initial experiments on sentiment analysis from GLUE \citep{wang2018glue} show that \method can successfully reduce the benchmark performance of several \llms, even without introducing the adversarial attack module.
More experimental results and analysis might come in the future and this paper shall be updated significantly.

\section{Related Work}
\label{sec-related}

Large language models achieve unprecedented performance across many tasks such as reasoning \citep{collins2023evaluating}, natural language processing \citep{parrish2021bbq}, and natural science applications \citep{guo2023indeed}.
There are several existing benchmarks to evaluate the performance of \llms such as AlpacaEval \citep{alpaca_eval}, OpenLLM leaderboard \citep{leaderboard}, Big-Bench \citep{srivastava2022beyond}, and API-bank \citep{li2023apibank}.
For a thorough overview of \llms evaluation, please refer to the survey paper \citep{chang2023survey}.

Of all the evaluation efforts to \llms, there are two main streams of work that share similar interests to ours.
One of them is the ``Dyna-X'' series, including DynaBoard \citep{ma2021dynaboard}, DynaTask \citep{thrush2022dynatask}, and DynaBench \citep{kiela2021dynabench}.
The key of these work is to leveraging the wisdom from the crowd for challenging evaluation sets design, i.e., the main efforts are not on the algorithm side, but in the crowd-sourcing system and interface, where the name ``dynamic'' comes from.
Our work is significantly different from theirs since we do not rely on crowd-sourcing for evaluation, but to generate OOD samples using our algorithm.

The other type of work is CheckList \citep{checklist} which automatically generates test samples for NLP datasets by replacing keywords in the given sentences.
Our work is similar to CheckList for the sentence generation part, but \method considers the semantic perseverance by using another language model acting as the evaluator LM.
On the other hand, the usage of another LM can help generate OOD samples (e.g., different styles or expressions with the same meaning), which CheckList might not have.
Therefore, our work can be seen as a more challenging version of CheckList in terms of measuring the semantic understanding ability of \llms.

Finally, red-teaming \citep{perez2022red} also uses templates for detecting toxic and offensive responses in language models. However, the templates in the Checklist and red-teaming include those that necessitate human manual effort for their creation, and the way of filling out a template is restricted to the task one template is used for. \method, on the other hand, can generate the templates automatically hence no human effort is needed.

\section{Methodology}
\label{sec-method}

In this section, we introduce our meta semantic template (\method) method for evaluating large language models.
As shown in \figurename~\ref{fig-main}, the core of \method is to generate semantically-preserved evaluation samples using different \llms.
In the following, we will introduce its key components: meta semantic templates generation (Sec. \ref{sec-method-semantic}) and real evaluation samples generation (Sec. \ref{sec-method-real}).
Then, we discuss the advantages and disadvantages of \method in Sec.~\ref{sec-method-discuss}.

\subsection{Meta Semantic Templates Generation}
\label{sec-method-semantic}

The process of meta semantic template generation includes both the blue and orange modules in \figurename~\ref{fig-main}.
Formally speaking, to evaluate an LLM $\mathcal{A}$, we need to have a seed dataset $\mathcal{D}=\{(x_i, y_i)\}_{i \in [N]}$, where $x$ is the input and $y$ is the output (ground truth label).
The generation of meta semantic template requires leveraging another language model, which we call \emph{evaluator LM}, $\mathcal{B}$, to generate $m$ semantics-preserving samples.
For an input $x_i$, e.g., ``\prompt{I am happy today}'' in \figurename~\ref{fig-main}, we generate $n$ samples, denoted as $\mathcal{S}_i=\{s_1, s_2, \ldots, s_n\}$, which are referred to as meta semantic samples, e.g., ``\prompt{Today brings me immense joy}'' and ``\prompt{My heart is filled with happiness on this day}''.

Notably, this step is non-trivial since we need to preserve the semantics of the generated sentences.
It requires efforts from two aspects.
First, we leverage an LLM $\mathcal{C}$ as the \emph{semantic preserving filter} to measure the similarity between the original and the generated sentences.
For instance, we choose $\mathcal{C}$ to be a BERT~\citep{devlin2018bert}-a powered filter that generates embedding for the original and generated sentences (denoted as $z$ and $z^\prime$, respectively).
Then, our filter computes their similarity score $c=\cos(z, z^\prime)$ and rank the scores for all generated embedding $z^\prime$.
There is a threshold $\tau$ to control how much of the generated sentences we want to preserve.
Second, the design of the prompts to $\mathcal{B}$ is non-trivial.
In order to generate sentences as different as possible, the prompts should be well-designed.
We have attempted different prompts like ``\prompt{Please generate 5 sentences with the same semantic meaning as the following sentence. Try different styles or expressions to make sure they are different.}''
This helps to generate different styles of sentences that suffice our needs.
However, the design of prompts is an open question that we believe should be improved in the future.

Then, for each $s_i$, we perform sentence parsing to generate the final meta semantic template $\mathcal{T}$.
This step is not related to any LLM since sentence parsing is relatively mature in the area of NLP.
We leverage existing libraries like NLTK \citep{bird2006nltk} to perform this operation.\footnote{Note that sentence parsing can also be performed using LLMs like ChatGPT, but we choose open-source libraries for efficiency and cost saving.}
This step generates meta semantic templates such as ``\prompt{[x] brings [x] immense joy}'' where ``\prompt{[x]}'' is replaceable to generate more real evaluation examples.

\begin{algorithm*}[t!]
    \caption{\method: Meta semantic template}
    \label{alg:mstemp}
    \begin{algorithmic}
        \STATE {\textbf{Input:}} seed dataset $\mathcal{D}$, an evaluator LM $\mathcal{B}$, semantic filter LM $\mathcal{C}$, threshold $\tau$
        \WHILE{not end}
            \STATE{Sample an example  $(x_i, y_i)$ in $\mathcal{D}$}
            \STATE{Generate $n$ meta semantic sentences $\mathcal{S}$ using $\mathcal{A}$ and $\mathcal{B}$ (satisfying $\tau \le \cos(\mathcal{B}(x_i), \mathcal{B}(s_j)$)}
            \STATE{Perform sentence parsing on $\mathcal{S}$ to get the meta semantic template $\mathcal{T}$}
            \STATE{Get real evaluation samples $\mathcal{D}^\prime$ by replacing words in meta semantic templates $\mathcal{T}$}
        \ENDWHILE
        \STATE \textbf{return} generated samples $\mathcal{D}^
        \prime$
    \end{algorithmic}
\end{algorithm*}

\subsection{Real Evaluation Samples Generation}
\label{sec-method-real}

This step is similar to CheckList \citep{checklist} which keeps generating different samples by randomly replacing words in the templates.
We denote the generated samples as $\mathcal{D}^\prime$.
For each template $s_i$, we generate $m$ samples such as ``\prompt{Bob brings Alice immense joy}'', where ``\prompt{Bob}'' and ``\prompt{Alice}'' are the new filled entities.
The replacement is conducted in a random manner to reduce the possibility of generating fixed samples.
The original ground truth labels (e.g., ``\prompt{sentiment: positive}'' in \figurename~\ref{fig-main}) are taken as the labels for the generated samples $\mathcal{D}^\prime$ thanks to the semantics preserving nature of the method.

Furthermore, the replacement can be made more challenging by introducing the adversarial attack module.
As shown in \figurename~\ref{fig-main}, the words ``\prompt{brigns}'' and ``\prompt{wit}'' are both typos generated by the adversarial attack module to fool the LLM.
And ``\prompt{joyful}'' is the outcome of the word-level attack generated as synonyms.
With the help of existing LLMs adversarial attack benchmarks \citep{zhu2023promptbench}, the generated samples can be more challenging to handle.
This makes \method more flexible.
The complete algorithm flow for \method is shown in Algo. \ref{alg:mstemp}.

\subsection{Discussion}
\label{sec-method-discuss}

\textbf{\method makes it possible to partially reduce the possibility of data contamination.}
In order to evaluate one LLM, we leverage $M$ extra evaluator LMs to generate different semantic templates for it.
Since different evaluator LMs could have different diversities and focus in a generation, we believe it could dramatically reduce the data contamination since a cheater might need huge efforts to collect these evaluator LMs and then generate samples.

On the other hand, the generation introduces randomness: the filter, the selected templates, and the replacement of templates all involve randomness.
In fact, \method makes it possible to generate different evaluation samples in its running every time.
This further makes it even more difficult to memorize the entire training data.

However, \textbf{\method has some limitations.}
First, there is some fairness issue in comparing the performance with existing benchmarks.
Imagine this: we leverage an $N$-size seed dataset $\mathcal{D}$ to generate $n$ templates for each input, and then we generate $m$ samples for each template.
This means we generate $nm$ real evaluation samples for \emph{each} input in $\mathcal{D}$: we will have $nmN$ evaluation samples in total by leveraging the full $\mathcal{D}$.
This makes it seem unfair to compare with the original performance in $\mathcal{D}$ since we literally have many more evaluation samples.
A potential solution to this is sampling: we sample $N$ examples from those $nmN$ generated samples.
We perform such sampling operations several times and use the average as the performance.

Another disadvantage of \method would be the guarantee of the naturalness and grammarly-corrected samples, which we do not control.
Our current control is through an extra LM and the filter to ensure that they are similar in embedding space, but this may not guarantee the similarity in the input level.
Additionally, the replacement of the words may introduce some grammar errors.
This limitation needs further solutions to handle.

\section{Experiments}

In this section, we conduct experiments to validate the effectiveness of \method.
We adopt SST-2 from GLUE \citep{glue} as the initial experiment.
We evaluate Flan-t5-Large \citep{wei2022finetuned} and Llama2-7b \citep{llama2} using \method.
For each LLM to evaluate, we choose two evaluator LMs: ChatGPT \citep{chatgpt} and Llama2-13b, to generate meta semantic templates.

\tablename~\ref{tb-main-results} shows our initial results.
It can be observed that after our evaluation, the performance on SST-2 is reduced.
Specifically, for Flan-T5-Large, the accuracy drops from $0.939$ to $0.877$ and $0.890$ using Llama2-13b and ChatGPT as evaluator LLMs, respectively.
For Llama2-7b, the accuracy drops from $0.813$ to $0.717$ and $0.709$, respectively.
The performance reductions for Flan-T5-Large and Llama2-7b are $\mathbf{5.9}\%$ and $\mathbf{12.3}\%$, respectively, if averaged on Llama2-13b and ChatGPT.\footnote{How to compute: $[0.939-(0.877+0.890)/2]/0.939\times100\%=5.9\%$.}
The results indicate that current \llms do have some limitations in correctly recognizing OOD samples.

\begin{table}[!ht]
\caption{Initial evaluation results using \method.}

\label{tb-main-results}
    \centering
    \resizebox{0.6\textwidth}{!}{
    \begin{tabular}{cccc}
    \toprule
\multicolumn{1}{l}{\multirow{2}{*}{Evaluated LLM}} & \multicolumn{1}{l}{\multirow{2}{*}{Baseline}} & \multicolumn{2}{l}{Evaluator LLM in \method} \\ \cline{3-4} 
\multicolumn{1}{l}{} & \multicolumn{1}{l}{} & Llama2-13b             & ChatGPT             \\ \midrule
        Flan-T5-Large & $0.939$& $0.877$& $0.890$\\
        Llama2-7b & $0.813$ & $0.717$ & $0.709$\\ 
        \midrule
        \#Examples & $872$ & $4081(4.68\times)$ & $4047(4.64 \times)$\\
        \bottomrule
        
    \end{tabular}
    }
    
\end{table}

\section{Conclusion}

This paper proposed \method, an evaluation approach to LLMs by generating semantically-preserving samples based on the given seed datasets.
\method has the potential to reduce the possibility of data contamination by involving OOD sample generation using extra evaluator LMs and replacement of words for templates.
We hope that this initial work can share some latest findings of our research towards LLMs evaluation and inspire new approaches in the future.

\section*{Disclaimer}

The generation mechanism of this work does not involve any irresponsible language.
The only purpose of conducting this research is to present a new evaluation protocol, but not to act as a new leaderboard to rank \llms.
Results using ChatGPT or other API services may not be reproducible due to API changes.

\bibliography{refs}
\bibliographystyle{plainnat}

\end{document}